\documentclass[nohyperref]{article}

\usepackage{microtype}
\usepackage{graphicx}
\usepackage{subfigure}
\usepackage{booktabs} 

\usepackage{hyperref}

\usepackage[accepted]{icml2023}

\usepackage{amsmath}
\usepackage{amssymb}
\usepackage{mathtools}
\usepackage{amsthm}

\usepackage[capitalize,noabbrev]{cleveref}

\usepackage{times}
\usepackage{latexsym}
\usepackage{amsmath}

\usepackage[T1]{fontenc}
\usepackage[utf8]{inputenc}

\usepackage{microtype}
\usepackage{bbding}
\usepackage{color,xcolor}
\usepackage{epsfig}
\usepackage{graphicx}

\usepackage{adjustbox}
\usepackage{array}
\usepackage{booktabs}
\usepackage{colortbl}
\usepackage{wrapfig}
\usepackage{hhline}
\usepackage{multirow}

\usepackage{amsmath,amsfonts,amssymb}
\usepackage{bm}
\usepackage{nicefrac}
\usepackage{microtype}
\usepackage{mathtools}

\usepackage{changepage}
\usepackage{extramarks}
\usepackage{fancyhdr}
\usepackage{lastpage}
\usepackage{setspace}
\usepackage{soul}
\usepackage{xspace}

\usepackage{url}

\usepackage{enumerate}
\usepackage{enumitem}  
\usepackage{titlesec}

\usepackage{makecell}

\usepackage{pifont} 

\renewcommand{\algorithmicrequire}{\textbf{Input:}}
\renewcommand{\algorithmicensure}{\textbf{Output:}}
\usepackage{amsthm}
\usepackage{float}

\usepackage{amsmath,amsfonts,bm}

\def\eqref#1{equation~\ref{#1}}

\def\1{\bm{1}}

\DeclareMathAlphabet{\mathsfit}{\encodingdefault}{\sfdefault}{m}{sl}
\SetMathAlphabet{\mathsfit}{bold}{\encodingdefault}{\sfdefault}{bx}{n}

\DeclareMathOperator*{\argmin}{arg\,min}

\newcommand{\fig}[1]{Figure~\ref{#1}}

\newcommand{\ignore}[1]{}

\renewcommand*{\thefootnote}{\fnsymbol{footnote}}

\DeclareMathAlphabet{\mathbfit}{OML}{cmm}{b}{it}

\makeatletter
\DeclareRobustCommand\onedot{\futurelet\@let@token\@onedot}
\def\@onedot{\ifx\@let@token.\else.\null\fi\xspace}

\def\eg{e.g\onedot} 
\def\ie{i.e\onedot}

\def\Method{Offsite-tuning\xspace}
\def\METHOD{Offsite-Tuning\xspace}
\def\method{offsite-tuning\xspace}

\newcommand{\myparagraph}[1]{\vspace{0pt}\paragraph{#1}}

\theoremstyle{plain}

\theoremstyle{definition}

\theoremstyle{remark}

\icmltitlerunning{\METHOD: Transfer Learning without Full Model}

\begin{document}

\twocolumn[
\icmltitle{\METHOD: Transfer Learning without Full Model}




\begin{icmlauthorlist}
\icmlauthor{Guangxuan Xiao}{mit}
\icmlauthor{Ji Lin}{mit}
\icmlauthor{Song Han}{mit}
\end{icmlauthorlist}

\icmlaffiliation{mit}{Massachusetts Institute of Technology}

\icmlcorrespondingauthor{Guangxuan Xiao}{xgx@mit.edu}

\icmlkeywords{Machine Learning, ICML}

\vskip 0.3in
]



\newcommand{\printAffiliations}[1]{%
\stepcounter{@affiliationcounter}%
{\let\thefootnote\relax\footnotetext{\hspace*{-\footnotesep}\ifdefined\isaccepted #1\fi%
\forloop{@affilnum}{1}{\value{@affilnum} < \value{@affiliationcounter}}{
\textsuperscript{\arabic{@affilnum}}\ifcsname @affilname\the@affilnum\endcsname%
\csname @affilname\the@affilnum\endcsname%
\else
{\bf AUTHORERR: Missing \textbackslash{}icmlaffiliation.}
\fi
}.
\ifdefined\icmlcorrespondingauthor@text
Correspondence to: \icmlcorrespondingauthor@text.
\else
{\bf AUTHORERR: Missing \textbackslash{}icmlcorrespondingauthor.}
\fi
}
}
}
\printAffiliations{}

\begin{abstract}
Transfer learning is important for foundation models to adapt to downstream tasks.
However, many foundation models are proprietary, so users must share their data with model owners to fine-tune the models, which is costly and raise privacy concerns. Moreover, fine-tuning large foundation models is computation-intensive and impractical for most downstream users.
In this paper, we propose \METHOD, a privacy-preserving and efficient transfer learning framework that can adapt billion-parameter foundation models to downstream data \emph{without} access to the full model.
In \method, the model owner sends a light-weight \textit{adapter} and a lossy compressed \textit{emulator} to the data owner, who then fine-tunes the adapter on the downstream data with the emulator's assistance.
The fine-tuned adapter is then returned to the model owner, who plugs it into the full model to create an adapted foundation model. 
\Method preserves both parties' privacy and is computationally more efficient than the existing fine-tuning methods that require access to the full model weights.
We demonstrate the effectiveness of \method on various large language and vision foundation models.
\Method can achieve comparable accuracy as full model fine-tuning while being privacy-preserving and efficient, achieving 6.5$\times$ speedup and 5.6$\times$ memory reduction. Code is available \href{https://github.com/mit-han-lab/offsite-tuning}{here}.
\end{abstract}
\section{Introduction}
\label{sec:intro}
\begin{figure}[t]
    \small
    \centering
    \includegraphics[width=\linewidth]{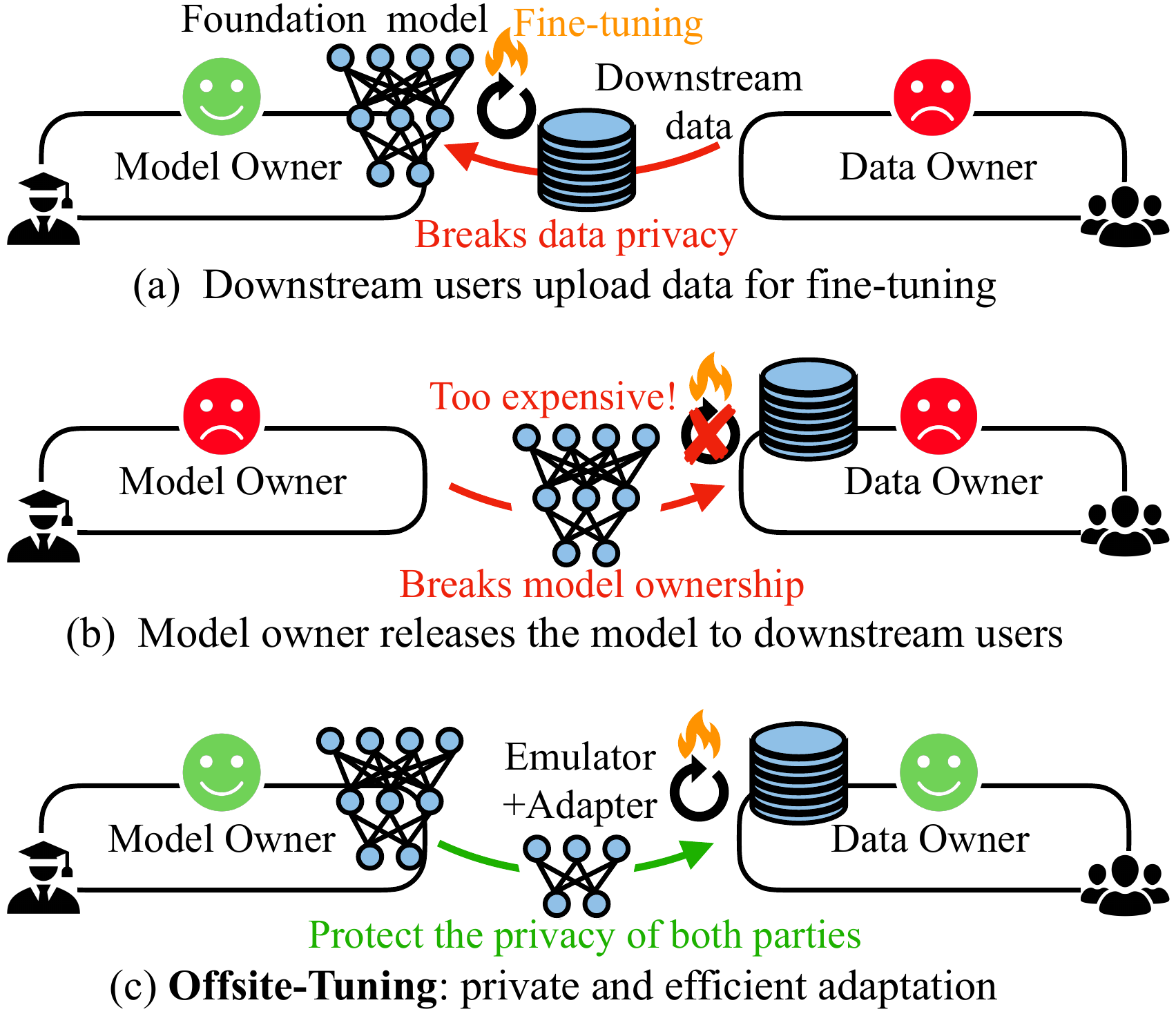}
    \caption{Comparing existing fine-tuning approaches (top and middle) and  \METHOD (bottom). (a) Traditionally, users send labeled data to model owners for fine-tuning, raising privacy concerns and incurring high computational costs. (b) Model owner sending the full model to the data owner is not practical, which threatens the ownership of the proprietary model, and it's not affordable for users to fine-tune the huge foundation model due to resource constraints. (c) \Method offers a privacy-preserving and efficient alternative to traditional fine-tuning methods that require access to full model weights.}
    \label{fig:paradigm}
\end{figure}
Large foundation models have shown exceptional performance in various tasks, including natural language processing~\cite{bert,gpt2,gpt3}, computer vision~\cite{clip,EVA}, and speech recognition~\cite{radford2022whisper}. Through pre-training on vast amounts of data, these models can learn general representations that are useful for a wide range of downstream tasks. Despite the ability of some foundation models to perform zero-shot predictions or in-context learning~\cite{gpt3,EVA}, transfer learning (\ie, fine-tuning) remains a popular and robust method to adapt a general foundation model to a specific task~\cite{ wei2021finetuned,instructgpt,bloomz,prefix,hu2021lora}.
\begin{figure}[t]
    \small
    \centering
    \includegraphics[width=\linewidth]{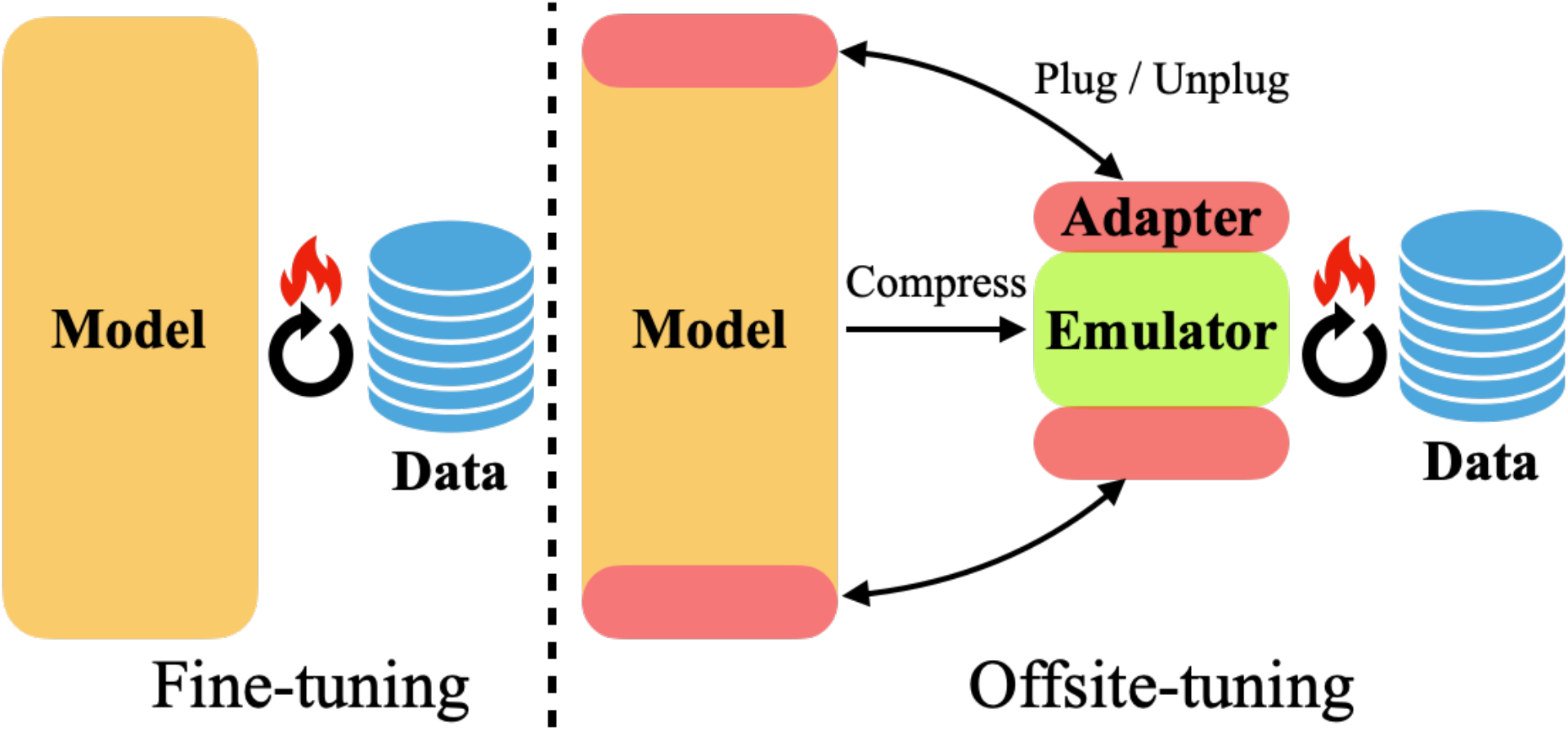}
    \caption{\textbf{Overview of \METHOD.} Fine-tuning (left) requires access to the full model weights and needs both model and data to be in one location. In \Method (right), the model owner sends an adapter and an emulator to the data owner, who fine-tunes the adapter on the downstream data with the emulator's assistance. The fine-tuned adapter is then returned and plugged into the full model to create an adapted foundation model. As neither party needs to share full models or data and the emulator is compressed, \method is both privacy-preserving and efficient.}
    \label{fig:overview}
\end{figure}
However, tuning foundation models for downstream tasks is difficult due to two reasons (Figure~\ref{fig:paradigm}). 
Firstly, training large foundation models usually require enormous computation and data, leading to high training costs (\eg, it is expected that it takes more than \$4M to train GPT-3\footnote{\url{https://lambdalabs.com/blog/demystifying-gpt-3}}). Therefore, the trained weights are usually proprietary and not made public. This means downstream users must share their labeled data with the model owners to fine-tune the models (\eg, the OpenAI Fine-tuning API\footnote{\url{https://beta.openai.com/docs/guides/fine-tuning}}), which can be costly and raise privacy concerns, putting valuable labeled data at risks. 
Secondly, even if the downstream users have access to the pre-trained weights, it is quite computationally expensive and difficult to perform the fine-tuning locally. Foundation models typically have a huge number of parameters. For example, the GPT-3 model~\cite{gpt3} has 175 billion parameters, requiring 350GB GPU memory to store the parameters and perform inference, let alone training. The demanding hardware requirement has made it impossible for most end users to perform transfer learning. 
Therefore, we need a privacy-preserving and more efficient framework for fine-tuning foundation models.

To address the challenges above, we propose \METHOD, a privacy-preserving and efficient transfer learning framework that can adapt foundation models to downstream tasks without access to the full model weights. As shown in \fig{fig:overview}, \method involves the model owner sending an \textit{adapter} and an \textit{emulator} to the data owner, who then fine-tunes the adapter on the downstream data with the emulator's assistance. The fine-tuned adapter is then returned to the model owner, who plugs it into the full model to create an adapted foundation model for downstream users. The adapters are used for encoding task-specific knowledge with a small number of parameters, while the compressed emulator mimics the behavior of the rest of the full model and provides approximate gradients for fine-tuning the adapters.
\Method preserves the \emph{privacy} of data owners, as they do not need to directly share their training data. It also protects the \emph{property} of the foundation model owner, as the full model weights are not shared, and the emulator is lossy compressed with highly degraded performance. \Method is also more \emph{resource-efficient} than existing fine-tuning methods that require access to the full model weights, as it allows fine-tuning without the need for the full model through the use of a compressed emulator.

We evaluate the performance of \method on a range of language and vision foundation models, including GPT-2, OPT, BLOOM, CLIP, and EVA. Results indicate that \method can achieve comparable results to fine-tuning with full model weights on multiple downstream tasks, while also preserving privacy and being more resource-efficient. Data owners can benefit from \method by adapting foundation models faster with less resource, achieving up to 6.6$\times$ speedup and 5.6$\times$ memory reduction compared to full fine-tuning. Additionally, \method enables fine-tuning of models that previously could not be achieved on a single GPU, such as OPT-6.7B and BLOOM-7.1B. Overall, we believe that \method is a practical framework for safely and efficiently applying foundation models to a broader range of real-world applications.

\section{Related Work}
\label{sec:related}
\textbf{Foundation Models}~\cite{bommasani2021opportunities}, also known as pre-trained models, are large neural networks that have been trained on a large dataset before being used for specific tasks. Although some models like GPT-3~\cite{gpt3}, CLIP~\cite{clip} and Painter~\cite{Painter} can make zero-shot predictions or learn in context, transfer learning remains a mainstream approach for applying models to new tasks~\cite{wei2021finetuned,bloomz,liu2022few}. Using foundation models can save time and resources compared to training models from scratch, but fine-tuning and deploying them can be resource-intensive due to their large parameter sizes~\cite{smith2022using,smoothquant}. Additionally, as many foundation models are non-public, users may need to share their training data with the model's owners for fine-tuning, which can be costly and raise privacy concerns.

\textbf{Parameter-Efficient Fine-tuning} adapts foundation models to downstream tasks by updating or adding only a small number of parameters, rather than updating the entire model. Techniques such as Adapter-tuning~\cite{adapter}, which inserts small task-specific neural networks into transformer layers, Prefix-tuning~\cite{prefix}, which prepends task-specific tunable prefix vectors to input sequences, LoRA~\cite{hu2021lora}, which decomposes the task-specific updates of weights into trainable low-rank vectors, and BitFit~\cite{bitfit}, which only updates the bias vectors of models, are useful as they require only a small number of parameters to be stored and loaded for each downstream task, while most parameters of the foundation model can be shared. However, it should be noted that while parameter-efficient fine-tuning is useful, it requires the knowledge of the entire model weights, which can compromise either data or model owners' privacy. Additionally, the fine-tuning process remains resource-intensive, as it requires at least one copy of the entire model to be placed on the device.

\textbf{Federated Learning}~\cite{mcmahan2017communication,konevcny2016federated,kairouz2021advances,augenstein2019generative} enables users to collectively train or fine-tune a model without sharing their data with a central server. Instead, each user maintains a local copy of the entire model and updates it with their data. The updated model is then sent to the central server, where it is aggregated to create a new global model. However, it is essential to note that while federated learning can protect data privacy by keeping the data on the devices, it does not preserve the model privacy as each user has a copy of the entire model. Furthermore, federated learning assumes that users can perform training on the whole model weights, which is hardly feasible for large foundation models.

\textbf{Decoupled Learning} breaks down the end-to-end optimization problem of neural network training into smaller sub-problems. This is achieved through various techniques such as the use of auxiliary variables~\cite{askari2018lifted,li2019lifted,taylor2016training,zhang2017convergent}, delayed gradient descent~\cite{huo2018decoupled,xu2020acceleration}, and model assembly~\cite{Ni2022Assemb}. However, current decoupled learning methods have primarily been developed for training neural networks from scratch and have not yet been extensively explored for fine-tuning already trained large foundation models.
\section{Problem Definition}
\label{sec:problem}


\textbf{Privacy requirements.}  We consider the privacy of two parties in the transfer learning setting:
the \emph{data owner} cannot share their labeled training data with the model owner, and the \emph{foundation model owner} cannot share their model with the data owner. 
We need to find a way to tune the model on the data owner's data without getting access to the full model weights. 

\textbf{Settings.}  Given the foundation model $\mathcal{M}$  parameterized by $\Theta$ and the downstream dataset $\mathcal{D}$, fine-tuning the model on the downstream datasets yields $\mathcal{M}_{\Theta} \rightarrow \mathcal{M}_{\Theta+\Delta}$, $\Delta = \argmin_{\delta} \mathcal{L}(\Theta+\delta, \mathcal{D})$. To enable private and efficient transfer learning, we want to find a substitute model $\mathcal{M}^*_{\Theta^*}$ (also called as \emph{Emulator}) that is (significantly) smaller and weaker than $\mathcal{M}_{\Theta}$, so that sharing $\mathcal{M}^*$ with downstream users would not threaten the ownership of the foundation models. 
Data owners then optimize the substitute model on the dataset, yielding $\mathcal{M}^*_{\Theta^*+\Delta^*}$. We hope that plugging the trained weights $\Delta^*$ back to the original model (\ie, $\mathcal{M}_{\Theta+\Delta^*}$) can achieve similar performance compared to directly optimizing $\mathcal{M}$ on the dataset (\ie, $\mathcal{M}_{\Theta+\Delta}$), without giving access to $\mathcal{M}$ itself.

\textbf{Metrics.}
To evaluate the performance of the method, we define several metrics. Without loss of generality, we use language models for the definitions.
\begin{itemize}
    \item \emph{Zero-shot performance} refers to the performance of the pre-trained foundation model when directly evaluated on downstream tasks \emph{without} fine-tuning (\ie, the performance of $\mathcal{M}_\Theta$). 
    \item \emph{Emulator performance} refers to the performance of the small substitute model when fine-tuned on the downstream datasets (\ie, the performance of $\mathcal{M}^*_{\Theta^* + \Delta^*})$ 
    \item \emph{Plug-in performance} refers to the performance of the pre-trained foundation model with plugged-in trained weights from the substitute model (\ie, $\mathcal{M}_{\Theta+\Delta^*}$).
    \item \emph{Full fine-tuning performance} refers to the performance when we \emph{directly} fine-tune the foundation model on downstream datasets (\ie, $\mathcal{M}_{\Theta+\Delta}$) without considering privacy. 
\end{itemize}

The core concept of \method is that downstream users can offsite fine-tune the foundation model on their private data without direct access to the full model. 
We accomplish this by generating emulated gradients with the emulator that can be leveraged to approximate update the adapters.
As such, we term our approach \METHOD.
 
 To prove the effectiveness of the method, we require:
 \begin{itemize}
     \item Zero-shot performance < plug-in performance, showing that the tuning effectively improves the performance on the specific downstream dataset (otherwise, tuning would not be necessary). 
     \item Emulator performance < plug-in performance, showing that the foundation model still edges in the task (otherwise, the downstream users would be happy to just use the fine-tuned emulator). 
     \item Plug-in performance $\approx$ full fine-tuning performance (so that users do not sacrifice too much performance for data privacy). 
 \end{itemize}

\section{\METHOD}
\label{sec:method}
\subsection{Framework Overview}
To attain the desired performance, we divide the foundation model, denoted as $\mathcal{M}$, into two distinct components: a small, trainable adapter, denoted as $\mathcal{A}$, which is intended for downstream adaptation, and the remaining portion of the model, denoted as $\mathcal{E}$, which is to be kept frozen. 
Specifically, $\mathcal{M}$ can be defined as the concatenation of $\mathcal{A}$ and $\mathcal{E}$, such that $\mathcal{M}=[\mathcal{A}, \mathcal{E}]$. 
To protect model ownership and improve efficiency, lossy compression is applied to the frozen component, resulting in an emulator, denoted as $\mathcal{E}^*$. 
The downstream user will be provided with the combination of the adapter and emulator, $[\mathcal{A}, \mathcal{E}^*]$, and will be able to perform model tuning by updating $\mathcal{A}$. 
The updated adapter, denoted as $\mathcal{A}'$, will then be returned to the upstream foundation model owner and integrated into the original model, $\mathcal{M}'=[\mathcal{A}', \mathcal{E}]$, to achieve superior performance on the downstream dataset. It is important to note that due to the lossy compression process, users with $[\mathcal{A}, \mathcal{E}^*]$ will not be able to achieve an acceptable level of performance, even with fine-tuning. Therefore, the integrity of the model ownership is not compromised during this process, and the overall efficiency is improved as a result of the compression.

However, determining an appropriate combination of $\mathcal{A, E, E^*}$ is a non-trivial task. Intuitively, the emulator, $\mathcal{E}^*$, should possess a level of similarity to the original frozen component, $\mathcal{E}$, to provide appropriate gradient directions for updating the adapter when fine-tuning on the downstream dataset. At the same time, the emulator cannot be too precise, as this would negate the need for the foundation models by the downstream user. In the following section, we will delve into the design of $\mathcal{A}, \mathcal{E}, \mathcal{E}^*$ and evaluate how different designs impact the aforementioned metrics.

\begin{algorithm}[tb]
    \caption{Uniform Layer-drop}
    \label{alg:layerdrop}
    \begin{algorithmic}
        \STATE {\bfseries Input:} a list of layers $[l_0, \ldots, l_{m-1}]$, number of retained layers $k$
        \STATE {\bfseries Output:} a list of retained layers $[l_{i_0}, \ldots, l_{i_{k-1}}]$

        \COMMENT{Make sure the first and last layers are retained}

        \STATE $stride \gets  (m-1) / (k-1) $ 
        
        \FOR{$j \gets 0$ to $k-1$}
            \STATE $i_{j} \gets \lfloor j \times stride \rceil$
        \ENDFOR
    \end{algorithmic}
    \vspace{-0.2em}
\end{algorithm}

\subsection{Adapter Selection}
The Transformer architecture~\cite{vaswani2017attention} has been widely adopted in foundation models across various modalities, such as language and vision. In this discussion, we will focus on the design of adapters for deep transformer backbones, which can easily be extended to other models, such as convolutional neural networks (CNNs).

We select a small subset of the foundation model as the adapter, which can be trained on various downstream datasets. Since only a subset of the model is updated, the adapter must be generalizable to different downstream tasks. Research has shown that different layers of a transformer, from shallow to deep, encode different levels of feature abstraction, and the selection of layers to update can impact transfer learning performance~\cite{lee2022surgical,lin2022device}. To cover a wide range of tasks, we choose to include both shallow and deep layers in the adapter, resulting in a sandwich design, $\mathcal{M} = \mathcal{A}_1 \circ \mathcal{E} \circ \mathcal{A}_2$. Our experiments show that this adapter design works well for various downstream tasks and outperforms the common practice of fine-tuning only the last several layers (\ie, $\mathcal{M} = \mathcal{E} \circ \mathcal{A}$), as demonstrated in Figure~\ref{fig:adapter_pos_num}. We include a detailed discussion of the adapter design in Section~\ref{sec:adapter_pos_num}.

\subsection{Emulator Compression}
The use of an emulator is to provide the rough gradient directions to update the adapters while remaining similar to the original frozen component, $\mathcal{E}$. However, the emulator must not be too precise, as this would reveal information about the original model. Additionally, a smaller emulator size leads to a more efficient fine-tuning process for downstream users. Therefore, we aim to find a balance between these three requirements.

To achieve this balance, we consider various compression methods, including pruning~\cite{han2016deep}, quantization~\cite{jacob2018quantization,smoothquant}, layer-drop~\cite{sajjad2023effect}, and knowledge distillation~\cite{hinton2015distilling,sanh2019distilbert}. Our experiments indicate that the layer-drop-based compression method provides the best balance between the aforementioned criteria. Specifically, we uniformly drop a subset of layers from the frozen component, $\mathcal{E}$, and use the remaining layers as the emulator, $\mathcal{E}^*$. We find it beneficial always to include the first and last layer of the frozen part in the emulator (as shown in Algorithm~\ref{alg:layerdrop}).
We provide a detailed comparison of different compression methods in Section~\ref{sec:emulator_compression}.
\begin{table*}[t]
\centering
\small
\caption{\Method (OT Plug-in) improves zero-shot (ZS) performance across all tasks, with only slight decreases compared to full fine-tuning (FT). Also, a consistent performance gap is observed between the emulator fine-tuning and plug-in, indicating \method effectively preserves the privacy of the original proprietary model (users can not use the emulator to achieve the same performance). }
\label{tab:lm_acc}
\begin{tabular}{lccccccccc}
\toprule
\textit{Setting} & OpenBookQA & PIQA    & ARC-E & ARC-C & HellaSwag & SciQ    & WebQs   & RACE    & WikiText ($\downarrow$) \\
\midrule\midrule
\multicolumn{10}{c}{GPT2-XL (2-16-2 Distill)} \\
Full ZS     & 23.0\%                 & 70.9\%                  & 58.2\%                  & 25.1\%                  & 40.0\%                  & 83.2\%                  & 1.5\%                   & 33.0\%                 & 20.44                 \\
Emulator ZS & 18.8\%                 & 67.7\%                  & 53.2\%                  & 20.8\%                  & 33.5\%                  & 77.0\%                  & 0.2\%                   & 30.0\%                 & 25.12                 \\
\midrule
FT          & 30.0\%                 & 73.2\%                  & 62.9\%                  & 30.0\%                  & 40.7\%                  & 92.5\%                  & 26.4\%                  & 43.2\%                 & 13.58                 \\
\midrule
OT Emulator &24.0\% & 70.3\% & 58.2\% & 23.9\% & 35.8\% & 92.7\% & 18.9\% & 39.4\% & 17.64  \\
OT Plug-in     &28.2\% & 73.6\% & 61.4\% & 28.5\% & 41.6\% & 93.2\% & 19.9\% & 39.9\% & 14.94  \\

\midrule\midrule
\multicolumn{10}{c}{OPT-1.3B (2-8-2 Distill)} \\
Full ZS     & 23.4\%                 & 71.6\%                  & 56.9\%                  & 23.5\%                  & 41.5\%                  & 84.4\%                  & 4.6\%                   & 34.2\%                 & 31.48                 \\
Emulator ZS & 19.4\%                 & 68.7\%                  & 53.9\%                  & 21.5\%                  & 35.1\%                  & 80.9\%                  & 1.3\%                   & 33.0\%                 & 38.55                 \\
\midrule
FT          & 31.4\%                 & 75.2\%                  & 61.3\%                  & 27.7\%                  & 42.7\%                  & 92.5\%                  & 31.2\%                  & 37.0\%                 & 12.52                 \\
\midrule
OT Emulator &24.8\% & 71.6\% & 58.1\% & 26.1\% & 37.0\% & 92.2\% & 24.3\% & 38.6\% & 15.54  \\
OT Plug-in     &29.0\% & 74.5\% & 59.4\% & 27.8\% & 43.3\% & 92.9\% & 26.2\% & 38.9\% & 13.15  \\

\bottomrule
\end{tabular}
\vspace{-1.5em}
\end{table*}

Additionally, to achieve a higher compression ratio while maintaining approximation accuracy, we apply knowledge distillation to the layer-dropped emulator, $\mathcal{E}^*$, under the supervision of the original component, $\mathcal{E}$, on the pre-training dataset when computing resources are available. The distillation process is performed using mean squared error (MSE) as the loss function, as shown in the following equation:
\begin{equation}
\mathcal{L}_{\text{distill}} = \frac{1}{N} \sum_{i=1}^N \Vert \mathcal{E}^*(x_i) - \mathcal{E}(x_i) \Vert^2
\end{equation}
where $x_i$ is the hidden representation of the $i$-th input sample produced by previous layers $\mathcal{A}_1$, $N$ is the number of samples in the pre-training dataset.  We find that a proper distillation will help maintain similar transfer learning accuracy compared to direct fine-tuning of the full model while still keeping an accuracy gap compared to the compressed model.
We discuss the effect of the distillation in Section~\ref{sec:emulator_distill}.
\subsection{Connection to Parameter-Efficient Fine-tuning}
Our work is orthogonal to existing parameter-efficient fine-tuning methods for foundation models, such as LoRA~\cite{hu2021lora}, Adapter~\cite{adapter}, BitFit~\cite{bitfit}, and Prefix-tuning~\cite{prefix}. Our focus is on fine-tuning a small subset of layers for data and model privacy while simultaneously improving efficiency through emulator compression. In contrast, parameter-efficient fine-tuning methods aim to reduce the number of trainable parameters while having access to the \emph{full} pre-trained weights. Our framework can easily integrate with these parameter-efficient fine-tuning methods by incorporating small trainable modules into the adapter component, $\mathcal{A}$. We will present the details of this integration and the results of our experiments in Section~\ref{sec:param_efficient}.
\begin{table*}[t]
    \centering
    \small
    \caption{\Method (OT Plug-in) improves zero-shot (ZS) performance across all tasks without the access to the full model. We observe a consistent performance gap between the emulator and plug-in performance, indicating \method effectively preserves the privacy of the original proprietary model.}
    \label{tab:llm_acc}
    \begin{tabular}{lccccccccc}
        \toprule
        \textit{Setting} & OpenBookQA & PIQA   & ARC-E  & ARC-C  & HellaSwag & SciQ   & WebQs  & RACE   & WikiText ($\downarrow$) \\
        \midrule\midrule
        \multicolumn{10}{c}{OPT-6.7B (2-18-2 Layer-drop)}                                                                         \\
        Full ZS          & 27.6\%     & 76.2\% & 65.6\% & 30.6\% & 50.5\%    & 90.1\% & 8.8\%  & 38.2\% & 24.24                   \\
        Emulator ZS      & 15.8\%     & 56.0\% & 33.8\% & 20.1\% & 28.3\%    & 51.1\% & 0.0\%  & 22.8\% & 44.92                   \\
        \midrule
        OT Emulator      & 23.4\%     & 59.4\% & 45.0\% & 19.5\% & 27.9\%    & 74.7\% & 8.0\%  & 26.7\% & 13.48                   \\
        OT Plug-in       & 33.8\%     & 77.7\% & 66.8\% & 33.9\% & 52.1\%    & 91.9\% & 23.9\% & 44.1\% & 10.78                   \\
        \midrule\midrule
        \multicolumn{10}{c}{BLOOM-7.1B (2-12-2 Layer-drop)}                                                                       \\
        Full ZS          & 24.8\%     & 72.7\% & 64.9\% & 30.3\% & 46.3\%    & 90.0\% & 2.3\%  & 36.6\% & 15.34                   \\
        Emulator ZS      & 16.4\%     & 64.0\% & 44.1\% & 20.1\% & 34.1\%    & 78.2\% & 0.1\%  & 28.2\% & 78.58                   \\
        \midrule
        OT Emulator      & 22.4\%     & 66.4\% & 52.9\% & 27.0\% & 35.9\%    & 85.7\% & 15.4\% & 30.3\% & 22.40                   \\
        OT Plug-in       & 29.6\%     & 74.6\% & 66.9\% & 36.8\% & 48.3\%    & 94.2\% & 25.0\% & 41.7\% & 14.58                   \\
        \bottomrule
    \end{tabular}
\end{table*}

\section{Experiments}
\label{sec:experiments}
\subsection{Setup}
\myparagraph{Models and Datasets.}
We evaluate \method on large language models, including GPT-2~\cite{gpt2}, OPT~\cite{opt}, and BLOOM~\cite{bloom}, as well as vision models such as CLIP~\cite{clip} and EVA~\cite{EVA}. 
We evaluate language models on the Wikitext language modeling dataset~\cite{wikitext} and eight question answering benchmarks, including OpenBookQA~\cite{OpenBookQA2018}, PIQA~\cite{piqa}, ARC~\cite{arc}, HellaSwag~\cite{hellaswag}, SciQ~\cite{SciQ}, WebQustions~\cite{webqs}, and RACE~\cite{race}. We report perplexity on the WikiText dataset and accuracy on the other benchmarks.
We evaluate vision models on six fine-grained image classification datasets: Flowers~\cite{flowers}, Cars~\cite{KrauseStarkDengFei-Fei_3DRR2013}, Pets~\cite{pets}, Food~\cite{food101}, CIFAR-10~\cite{Krizhevsky09learningmultiple}, and CIFAR-100~\cite{Krizhevsky09learningmultiple}. The performance of these models is measured in terms of accuracy.
\myparagraph{Implementation Details.}
Our experiments are based on the Huggingface transformers library\footnote{\url{https://github.com/huggingface/transformers}}.  During training and fine-tuning, we employ the AdamW~\cite{adam} optimizer and a cosine learning rate scheduler. We tune the learning rate on a grid of values: \{2e-5, 5e-5, 1e-4, 2e-4, 3e-4\}, and report the runs with the highest emulator performance. This is because, in real-world scenarios, users will return the adapter with the best Emulator performance to the model owner.
We use {\texttt{lm-eval-harness}}\footnote{\url{https://github.com/EleutherAI/lm-evaluation-harness}} for language model evaluation and NVIDIA A6000 GPUs for all experiments.
\subsection{Results of Language Models} 
\myparagraph{Medium-sized models.}
We first evaluate \method on medium-sized language models with less than 2 billion parameters, including GPT-2-XL~\cite{gpt2} and OPT-1.3B~\cite{opt}. Specifically, GPT-2-XL has 48 layers and 1.6 billion parameters, and OPT-1.3B has 24 layers.
On the medium-sized language models (less than 2 billion parameters), we have the compute resource to fine-tune the full model and use knowledge distillation to compress the emulator. We use the top and bottom two transformer layers of the model as the adapter. We reduced the number of layers to 16 and 8 for GPT-2-XL and OPT-1.3B, respectively, as the initialization for emulator distillation. Next, we distilled the emulator on the first of 30 chunks of the Pile corpus training set for a single epoch. The results are reported in Table~\ref{tab:lm_acc}. We find that \method can effectively adapt medium-sized language models while maintaining a high level of performance as the plug-in performance is comparable to full model fine-tuning performance, while the emulator performance is significantly lower.
\myparagraph{Large models.}
We then evaluate \method on large language models with more than 6 billion parameters, including OPT-6.7B~\cite{opt} with 32 layers and BLOOM-7.1B~\cite{bloom} with 30 layers. Due to the limited computational resources, we are \emph{unable} to perform full model fine-tuning or emulator distillation on these models (model owner should be able to perform the distillation, which costs a small fraction compared to pre-training). As a result, we compare the performance of \method with the zero-shot performance and directly use the layer-drop method to get the emulator. We use the top and bottom two transformer layers of the model as the adapter. We dropped the middle 28 and 16 layers to 18 layers and 12 layers for OPT-6.7B and BLOOM-7.1B, respectively. The results are reported in Table~\ref{tab:llm_acc}. From the results, we find that the plug-in performance is significantly better than the zero-shot performance, while there is a noticeable gap between the emulator performance and the plug-in performance. These findings indicate that \method can effectively adapt large language models while preserving the privacy of the model and the data owners.
\begin{table}
    \setlength{\tabcolsep}{2.5pt}
    \centering
    \small
    \caption{\Method also works on vision foundation models. Scores are percent accuracy.}
    \label{tab:vit_acc}
    \begin{tabular}{lcccccc}
    \toprule
    \textit{Setting} & Flowers                     & Cars    & Food    & Pets    & CF-10 & CF-100  \\
    \midrule\midrule
    \multicolumn{7}{c}{EVA-ViT-G (4-22-4 Distill)} \\ 
    FT                      & 99.59                     & 95.77 & 95.43 & 95.80 & 99.39 & 93.96   \\
    \midrule
    OT Emulator                    & 99.15                     & 94.58 & 94.90 & 96.24 & 99.52 & 93.31   \\
    OT Plug-in                     & 99.33                     & 94.99 & 95.33 & 96.48 & 99.63 & 94.21   \\
    \midrule\midrule
    \multicolumn{7}{c}{CLIP-ViT-G (4-16-4 Distill)} \\
    FT            & 99.33                     & 95.65 & 94.29 & 94.99 & 99.13 & 92.33   \\
    \midrule
    OT Emulator        &  99.01& 95.10 & 92.91 & 95.26 & 99.11 & 90.47   \\
    OT Plug-in                     & 99.27                     & 95.16 & 93.88 & 95.53 & 99.13 & 91.00  \\
    \bottomrule
    \end{tabular}
\vspace{-1em}
\end{table}
\subsection{Results of Vision Models}
We further evaluate \method on two state-of-the-art vision foundation models: CLIP~\cite{clip} and EVA~\cite{EVA}. Both models utilize ViT-G backbones with 1 billion parameters. We use the OpenCLIP~\cite{openclip} checkpoint trained on LAION-2B~\cite{laion}\footnote{\url{https://huggingface.co/laion/CLIP-ViT-g-14-laion2B-s12B-b42K}} for CLIP and the EVA-CLIP checkpoint trained on LAION-400M~\cite{laion400m}\footnote{\url{https://huggingface.co/BAAI/EVA/blob/main/eva_clip_vis_enc_sz224_lincls_86p5.pth}} for EVA. We use the top and bottom four transformer layers plus the classification head as the adapter. To initialize the emulator, we reduce the number of layers to 16 and 22 for CLIP and EVA, respectively. We then distill the emulator on ImageNet~\cite{imagenet} for a single epoch. As shown in Table~\ref{tab:vit_acc}, we find that \method effectively adapts the vision models while maintaining a high level of performance. The plug-in performance is comparable to full model fine-tuning performance, while the emulator performance is slightly lower. This may be due to the fact that the difference between using large and small vision models on these datasets is not significant, and thus the emulator performance is not significantly lower than the plug-in performance. We anticipate that \method will yield more significant results on more challenging vision tasks.
\begin{figure}[t]
    \small
    \centering
    \includegraphics[width=0.95\linewidth]{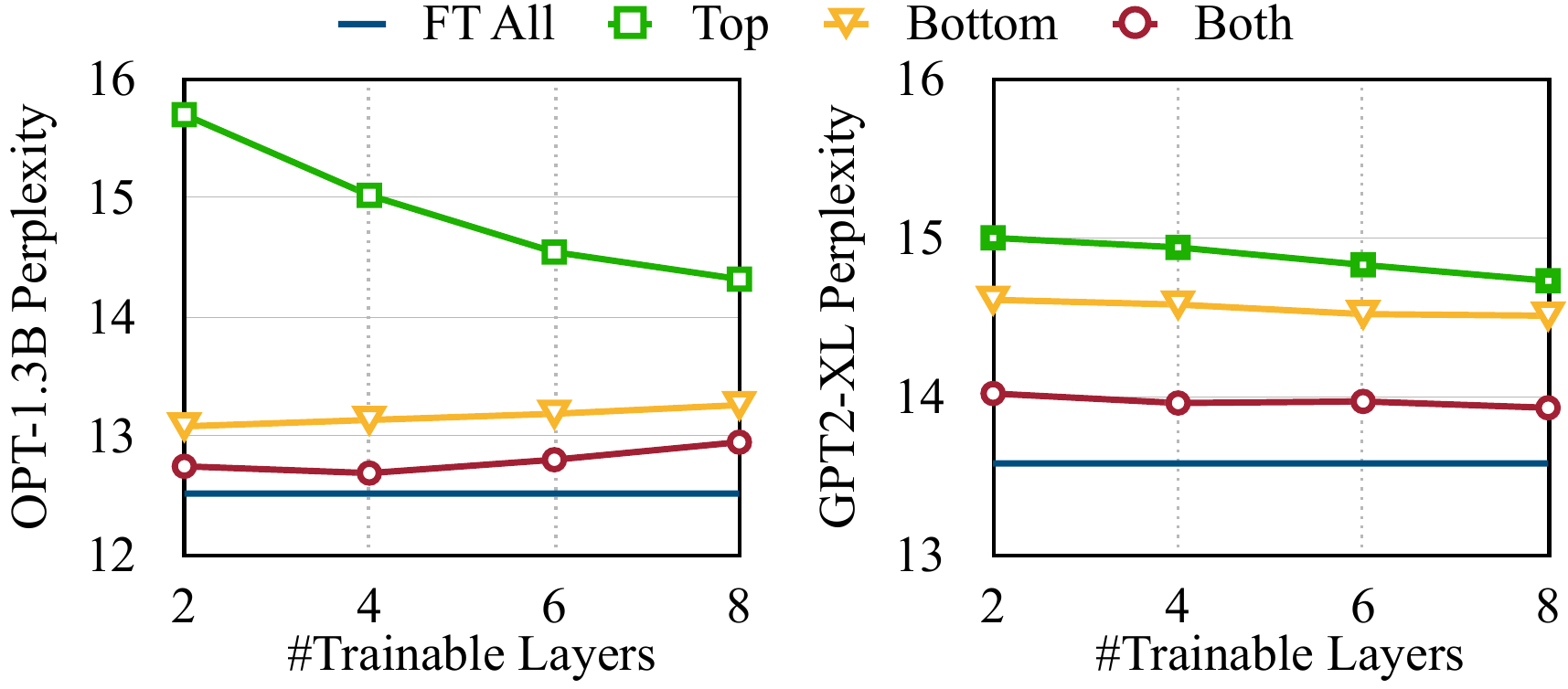}
    \caption{Ablation study of the number and position of adapter layers. Fine-tuning both the top and bottom layers of the language model is significantly more effective than fine-tuning only the top or bottom layers, given the same number of trainable layers.}
    \label{fig:adapter_pos_num}
\end{figure}
\begin{figure}[t]
    \small
    \centering
    \includegraphics[width=\linewidth]{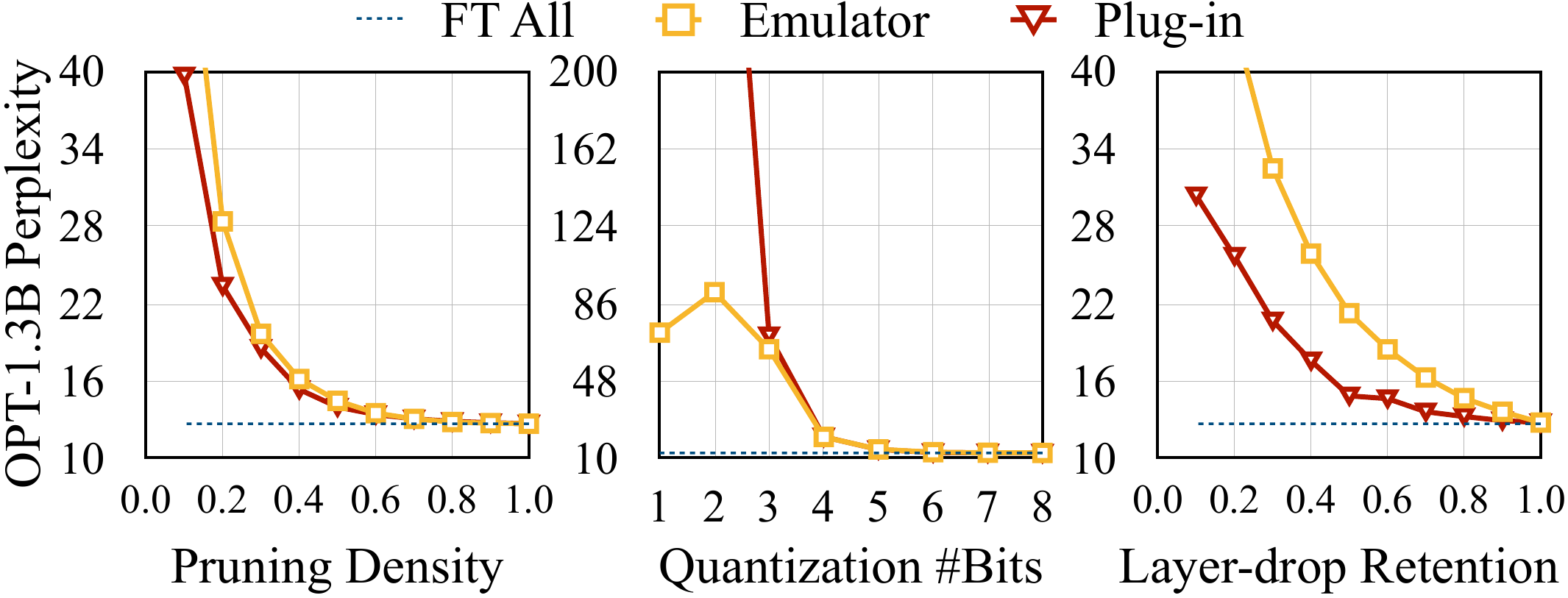}
    \caption{Ablation study of compression methods for creating the emulator. The layer-drop method is superior in two aspects: (1) it effectively maintains the plug-in performance while reducing the size of the emulator; (2) it creates a gap between the plug-in performance and the emulator performance, preserving the privacy of the model owner.}
    \label{fig:emulator_compression}
\end{figure}
\begin{table}[t]
    \small
    \setlength{\tabcolsep}{2.5pt}
    \centering
    \caption{Distilling the emulator improves \method's plug-in performance while maintaining full model privacy.  Scores are validation perplexities on WikiText-2 (lower is better).}
    \label{tab:distill}
    \begin{tabular}{lcccc}
        \toprule
                   & \multicolumn{2}{c}{w/o Distillation}                           & \multicolumn{2}{c}{w/ Distillation}                                       \\
                   \cmidrule{2-3} \cmidrule{4-5}
                   & Emulator                 & Plug-in                    & Emulator                 & Plug-in                     \\
                   \midrule
   GPT2-XL (2-16-2) & 30.48 & 19.69 & 17.64 & 14.94 \\
    OPT-1.3B (2-8-2) & 18.44 & 14.63 & 15.54 & 13.15  \\

    \bottomrule
    \end{tabular}
    \vspace{-1em}
\end{table}
\subsection{Ablation Study}
\subsubsection{Position and Number of Adapter Layers}
\label{sec:adapter_pos_num}
In Figure~\ref{fig:adapter_pos_num}, we show the effect of the position and number of adapter layers on the fine-tuning performance. We compare the performance of only tuning the top layers (near the output), bottom layers (near the input), and evenly both top and bottom layers while freezing the rest of the layers. We find that given the same budget of trainable layers, evenly tuning both the top and bottom layers significantly outperforms tuning only the top or bottom layers. Additionally, the performance gap between partial fine-tuning and full fine-tuning is not significantly reduced when we increase the number of tunable layers. Based on these results, we chose to use the top and bottom two layers as the adapter throughout our experiments.
\subsubsection{Compression methods for the Emulator}
\label{sec:emulator_compression}
In addition to the layer-drop and distillation methods that we primarily applied in our experiments, we also investigate other compression techniques to create an emulator, including magnitude-based pruning~\cite{han2016deep} and quantization~\cite{jacob2018quantization,smoothquant}. To construct the emulator, we use two bottom and top layers as the adapter and compress the middle layers with various compression methods.
Our results, presented in Figure~\ref{fig:emulator_compression}, show that layer-drop achieves the best performance when plugging into the original model. Furthermore, we observe a clear gap between the plug-in and emulator performance when using the layer-drop method, while this gap is not significant when using other compression methods. This suggests that the layer-drop-based compression method for creating the emulator can effectively protect the model's privacy while maintaining a high level of plug-in performance. Overall, our results demonstrate that the layer-drop-based method is an effective approach for creating efficient and privacy-preserving emulators.
\subsubsection{Effect of Emulator Distillation}
\label{sec:emulator_distill}
In Table~\ref{tab:distill}, we show that the emulator distillation can further improve the plug-in performance. Specifically, with the same number of emulator layers, the plug-in performance is improved by 2.47 on the OPT-1.3B model and 4.75 on the GPT-2-XL model. Despite these improvements, we still observe a clear gap between the plug-in and emulator performance, so the privacy of the full model is still well preserved. This suggests that there is further potential to improve the parameter efficiency of the emulator if we have access to more compute resources to perform additional distillation epochs.
\begin{table}[t]
    \setlength{\tabcolsep}{2.5pt}
    \small
    \centering
    \caption{\Method is orthogonal to and can be combined with parameter-efficient fine-tuning techniques. Scores are validation perplexities on WikiText-2 (lower is better).}
\label{tab:param_efficient}
    \begin{tabular}{lrcc}
    \toprule
      & \shortstack{\#Trainable\\Parameters} & \multicolumn{1}{l}{Emulator PPL} & \multicolumn{1}{l}{Plug-in PPL}  \\
\midrule\midrule
\multicolumn{4}{c}{GPT2-XL (2-16-2 Distill)} \\
FT           & 1475M                  & -                               & 13.58                           \\
\midrule
OT       & 123M                   & 17.64                           & 14.94                           \\
\quad + Adapter & 1.65M                  & 17.88                           & 14.99                           \\
\quad + LoRA    & 410K                   & 18.42                           & 14.84                           \\
\quad + BitFit  & 83K                    & 21.96                           & 18.75                           \\
\midrule\midrule
\multicolumn{4}{c}{OPT-1.3B (2-8-2 Distill)} \\
        FT           & 1208M                  & -                               & 12.52                           \\
        \midrule
OT       & 201M                   & 15.55                           & 13.15                           \\
\quad + Adapter & 2.11M                  & 15.93                           & 13.40                           \\
\quad + LoRA    & 590K                   & 15.66                           & 13.20                           \\
\quad + BitFit  & 106K                   & 18.53                           & 17.88                           \\
\bottomrule
\end{tabular}
\end{table}
\subsection{Combining with Parameter-Efficient Fine-tuning}
\label{sec:param_efficient}
\Method is orthogonal and can be seamlessly combined with existing parameter-efficient fine-tuning methods to further reduce the number of trainable parameters for each task. To combine \method with parameter-efficient fine-tuning, we only need to apply the method on the adapter layers. We use Adapter-tuning~\cite{adapter}, LoRA~\cite{hu2021lora}, and BitFit~\cite{bitfit} as examples to demonstrate the effectiveness of this combination. We set the adapter size to 64 for Adapter-tuning and rank 4 for LoRA. We perform our experiments on the OPT-1.3B and GPT-2-XL models on the WikiText dataset. As shown in Table~\ref{tab:param_efficient}, we find that Adapter-tuning and LoRA can significantly reduce the number of trainable parameters while maintaining the plug-in performance. However, we also observe that BitFit failed to adapt the model as there is a significant gap between the plug-in performance and the full model fine-tuning performance.
\begin{table}
    \small
    \centering
    \caption{Fine-tuning speedup and peak memory saving of \method on a single A6000 GPU. Throughputs are in tokens per second and memory is in megabytes (MB). Batch sizes are 1 and sequence lengths are 512.}
    \label{tab:efficiency}
    \begin{tabular}{llcc}
        \toprule
                                  & Method      & Throughput ($\uparrow$) & Memory ($\downarrow$) \\
                \midrule
        \multirow{4}{*}{GPT2-XL}  & FT          & 957                     & 39922                 \\
                                  & + LoRA      & 1992                    & 17113                 \\
        \cmidrule{2-4}
                                  & OT (2-16-2) & 4905                    & 8966                  \\
                                  & + LoRA      & \textbf{6257}           & \textbf{7155}         \\
        \midrule
        \multirow{4}{*}{OPT-1.3B} & FT          & 1316                    & 30458                 \\
                                  & + LoRA      & 3379                    & 12470                 \\
        \cmidrule{2-4}
                                  & OT (2-8-2)  & 5207                    & 9336                  \\
                                  & + LoRA      & \textbf{8525}           & \textbf{6192}         \\
        \bottomrule
    \end{tabular}
    \vspace{-1em}
\end{table}
\subsection{Efficiency}
The key efficiency advantage of \method is its ability to reduce not only the number of trainable parameters but also the total number of parameters that need to be placed on the device during fine-tuning. This results in a significant increase in fine-tuning throughput and a reduction in memory footprint. To demonstrate the effectiveness of \method, we conduct experiments and present the results in Table~\ref{tab:efficiency}. The results show that when \method is combined with LoRA, we achieve an impressive 6.5x speedup and a 5.6x reduction in memory usage. This makes \method an attractive solution for fine-tuning large foundation models on resource-constrained devices.

\section{Discussion}
\label{sec:discussion}
\myparagraph{Use cases.}
\Method is an effective approach for personalizing large language models on edge devices, making it useful for various applications such as voice assistants and chatbots. For example, users can utilize \method to adapt a large language model to their personal information directly on their devices, which is more efficient and preserves privacy by eliminating the need to send data to a server. The adapted model can then be used to generate personalized text, such as emails and messages. Additionally, \method can be applied in domains where training data is extremely sensitive and cannot be shared, such as in a hospital setting where it can be used to adapt a large language model to patient records without sharing the records with the model owner. The adapted model can then be used to generate personalized medical reports for patients. In both cases above, the method also protects the privacy of the model owner, as they do not need to share their full model with the data owner.
\myparagraph{Inference privacy.}
In this work, we focus on addressing the data privacy and efficiency concerns during the process of adapting or fine-tuning foundation models, rather than addressing privacy issues during inference. The proprietary training datasets for downstream tasks are often well-labeled and contain significant business values, making privacy a crucial consideration. However, privacy concerns during inference can be addressed through other methods such as~\cite{chou2018faster,li2022mpcformer}.
\myparagraph{Limitation and future works.}
While our results indicate that using approximately one-third of the middle layers of a foundation model as an emulator is viable, it remains huge for models such as GPT-3. Also, compression of the emulator through compute-intensive distillation techniques may be cost-prohibitive for larger models. Furthermore, we have yet to fully demonstrate that our method does not inadvertently result in model and data information leakage. Future research should investigate the possibility of reconstructing the full model and downstream data from the emulator and adapter. Additionally, the theoretical foundation of our method remains unclear and further investigation is needed to provide insights into the emulator and adapter design.

\section{Conclusion}
\label{sec:conclusion}
We propose \method, a privacy-preserving and efficient transfer learning framework that can adapt foundation models to downstream tasks without access to full model parameters. \Method is effective on billion-parameter language and vision foundation models. \Method enables  users to efficiently customize foundation models without worrying about data privacy and model privacy.


\section*{Acknowledgements}
We thank MIT-IBM Watson AI Lab,  
MIT AI Hardware Program,
Amazon and MIT Science Hub, 
NVIDIA Academic Partnership Award,  
Microsoft Turing Academic Program,
Qualcomm Innovation Fellowship, 
and NSF for supporting this research. We thank Tianwei Yin for the helpful discussions.


\bibliography{references}
\bibliographystyle{icml2023}

\end{document}